# Decentralized Sensor Fusion with Distributed Particle Filters

Matt Rosencrantz, Geoffrey Gordon, and Sebastian Thrun
School of Computer Science
Carnegie Mellon University
Pittsburgh, PA 15213

## Abstract

This paper presents a scalable Bayesian technique for decentralized state estimation from multiple platforms in dynamic environments. As has long been recognized, centralized architectures impose severe scaling limitations for distributed systems due to the enormous communication overheads. We propose a strictly decentralized approach in which only nearby platforms exchange information. They do so through an interactive communication protocol aimed at maximizing information flow. Our approach is evaluated in the context of a distributed surveillance scenario that arises in a robotic system for playing the game of laser tag. Our results, both from simulation and using physical robots, illustrate an unprecedented scaling capability to large teams of vehicles.

## 1 Introduction

In recent years, probabilistic tracking techniques have found widespread application in computer vision (Isard & Blake, 1998), diagnostics (Lerner et al., 2002; Verma et al., 2003), modeling (Rusinkiewicz & Levoy, 2001), intelligent environments (Pasula et al., 1999), and robotics (Thrun, 2002). Virtually all state-of-the-art approaches are based on the basic Bayes filter (Jazwinsky, 1970; Maybeck, 1990). While Bayes filters (and derivatives) are easily defined for centralized architectures, in which a single computational node receives all sensor data, they have proven difficult to extend to distributed systems.

In distributed systems sensor data is acquired through multiple platforms and the communication between these platforms is the major bottleneck. Examples of distributed systems include agents on the Internet, environments instrumented with many sensors, multi-camera surveillance systems, and teams of mobile robots pursuing a joint goal.

The simplest kind of architecture for state estimation is *centralized*. In centralized architectures, all platforms communicate all sensor data to a single special agent, which processes it centrally and broadcasts the resulting state estimate back to the individual platforms. Such an approach suffers from two problems: First, it is brittle because it possesses a single point of failure. Second, the enormous communication overhead often prohibits its use in situations with more than a few dozen platforms. Both limitations have long been recognized in the decentralized tracking literature (Rao et al., 1993).

*Decentralized* architectures, by contrast, rely on communication between nearby platforms. By doing so, the number of messages that each platform sends or receives is *independent* of the total number of platforms in the system. This property ensures scalability to distributed systems with (almost) any number of platforms.

In a recent result (Nettleton et al., 2000), it was shown that scalable decentralized state estimation can be achieved for *static environments with Gaussian error*—that is, environments where the state does not change over time and observations are (approximately) linear functions of the state with Gaussian error. The key insight in (Nettleton et al., 2000) is that in static worlds, information can be accumulated at any time and in any order, regardless of the time at which it was acquired. Furthermore, because of the simple form of the observations, there is a simple algorithm for accumulating local evidence in the form of additive information matrices. Unfortunately, the "trick" of additive information works only for distributions (such as Gaussians) which are in the exponential family. Attempts to apply similar ideas to more general observations tend to lead to overconfident state estimates (Fox et al., 2000). More importantly, such approaches are inapplicable to dynamic environments, since we are no longer free to incorporate information in any order.

This paper extends previous research in two critical directions: our decentralized filter can cope with dynamic environments and it can handle non-Gaussian posteriors. As in (Nettleton et al., 2000), each platform maintains its own Bayesian state estimate, computed from its local sensor measurements and information received from other (nearby) platforms. In contrast to (Nettleton et al., 2000), however, each platform maintains a belief over state histories instead of just single states. As we shall see below, this extension enables our approach to cope with dynamic environments. However, keeping this extra information introduces a new problem: because the space of possible state histories is so large, it is no longer feasible to represent, communicate, or incorporate evidence about full histories directly.

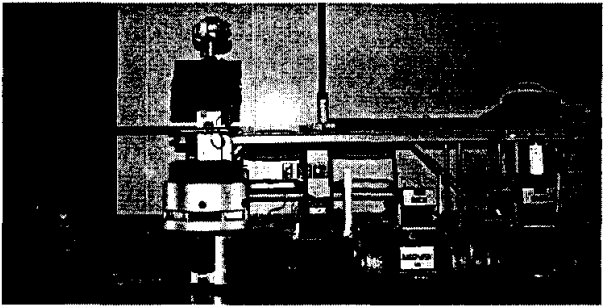

Figure 1: The CMU laser tag system consists of six robots based on Pioneer and Scout platforms, of which five are shown here. The robots are equipped with infrared emitters and detectors. Navigation is achieved using the CARMEN software system (Montemerlo et al., 2002).

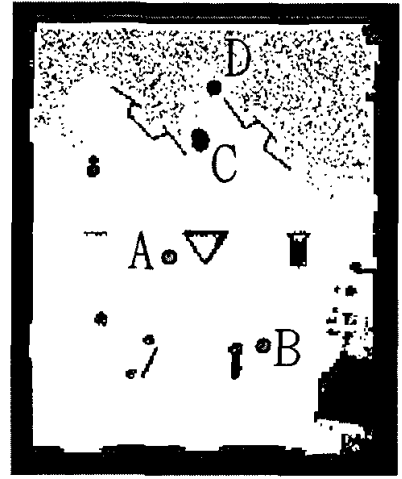

Figure 2: A typical belief in robot laser tag. The small circles are robots. This belief is from the perspective of robot **A** who is sharing information with its teammate **B**. Together, they are tracking opponents **C** and **D**. The belief **D**'s position is highly non-Gaussian, as illustrated by the particles.

The key innovation of our approach is a selective communication scheme which never attempts to send evidence about full histories around the network. Our algorithm takes the form of a query-response protocol. Platforms query their neighbors by sending compact summaries of their posterior beliefs. The neighbors keep databases of past sensor measurements, and respond by sending the *most informative* piece of evidence in their local memory. So, information exchange is efficient, and our communication overhead can be kept to a minimum.

Our approach has been implemented for a distributed surveillance scenario (Dolan et al., 1999) motivated by ongoing research on a robotic laser tag system (Rosencrantz et al., 2003). This domain involves two teams of robots pursuing each other in an indoor environment, tagging each other using infrared emitters (see Figure 1). The domain is characterized by massive occlusion, as opponents are usually hidden behind obstacles. This naturally leads to non-Gaussian, multi-modal beliefs, as illustrated in Figure 2. So, our implementation uses particle filters (Doucet et al., 2001; Liu & Chen, 1998) to estimate non-Gaussian posteriors over state sequences.

Recently there has been some other efforts to design distributed Bayes filters in the sensor networks community (Chu et al., 2002). However, this work focuses on the problem of querying a network of sensors to provide an answer for an external party, whereas we are interested in scenarios where the sensors are also the consumers of information, in particular where each needs to maintain an accurate posterior. Because of this their technique does not provide a mechanism to disseminate relevant information to all sensor nodes. Nor does it allow new sensors to join the network and contribute observations that were collected at some point in the past, whereas our algorithm is specifically designed for such situations. There has also been some work toward deriving general minimum distortion bounds for certain topologies of sensor networks (Chen et al., 2001), in contrast we will exploit features of our domain to simplify our problem.

Our approach extends previous work on multi-robot tracking (Rosencrantz et al., 2003) (which required a centralized tracker) to a fully decentralized system. As a result, our approach performs well using 50 robots (in simulation); in fact, it consistently outperforms alternative decentralized algorithms. To the best of our knowledge, such scalability is unprecedented in a dynamic environment. We also provide results obtained using our physical robot system, albeit with many fewer robots per team.

## 2 The Decentralized Bayes Filter

### 2.1 Static Environments

To goal of Bayesian state estimation is to calculate a posterior distribution over the state. Let us denote state by $x$. Naturally, the state is not directly observable but has to be inferred from data. Let us denote the number of platforms (robots) by $N$. The data acquired by the $i$-th platform at time $t$ will be denoted $d_t^i$; here the subscript refers to the time and the superscript to the platform identity. The set of all data items collected between the initial time, denoted 0, and time $t$, will be collectively denoted by $d_{0:t}^i$. The desired posterior is given by

$$p(x \mid d_{0:t}^1, \ldots, d_{0:t}^N) \qquad (1)$$

The posterior can be conveniently calculated by combining local posteriors, obtained by integrating the sensor data acquired by a local platform. To see, we note that

$$\begin{aligned} p(x \mid d_{0:t}^1, \ldots, d_{0:t}^N) &\propto p(x) \, p(d_{0:t}^1, \ldots, d_{0:t}^N \mid x) \\ &= p(x) \prod_{i=1}^N p(d_{0:t}^i \mid x) \\ &\propto p(x) \prod_{i=1}^N \frac{p(x \mid d_{0:t}^i)}{p(x)} \end{aligned} \qquad (2)$$

Here $p(x \mid d_{0:t}^i)$ is the local posterior of the $i$-th platform, and $p(x)$ is simply the prior over the state. The ratio of the two is the *evidence* accumulated by the $i$th platform about the state $x$.

For beliefs from the exponential family (e.g., Gaussians), each agent's evidence can be calculated in closed form and can be represented by a message of fixed size no matter how many observations the agent has seen. Furthermore, the product in Eq. (2) can be calculated in closed form and in arbitrary order: when we find out the evidence from each agent we can just multiply it into a running total (or add it in if we are accumulating on a log scale).

These insights make possible several important computational simplifications: if an agent updates its evidence all we need to do is divide out the old evidence and multiply in the new; and if we have evidence available from several platforms we can combine it all into a single message before passing it to our neighbors. In this way, a globally consistent result can be achieved using a communication structure where only neighbors communicate, and where the number of messages per platform is independent of the number of platforms. This observation forms the core of a literature on decentralized sensor fusion, of which (Nettleton et al., 2000) is an example.

### 2.2 Dynamic Environments

Unfortunately, extending this approach to dynamic environments is *not* straightforward. In dynamic environments, the state can change over time; so, we must annotate the state with a time index. We will write $x_t$ for the state at time $t$, and $x_{0:t}$ for the sequence of all states from time 0 to $t$. Unfortunately, the factorization in Eq. (2) does not apply if the static state $x$ is replaced by the dynamic state $x_t$. To see why, consider the example in Figure 3. This figure illustrates a distributed surveillance problem, with two platforms indicated by circles with direction markers. Suppose that the opponent is known to start off in the shaded region of Figure 3a. Because the opponent can move, at some point in time $t$ the local posterior $p(x \mid d_{0:t}^i)$ of the top robot is illustrated by the shaded region in Figure 3b. The posterior of the other robot is symmetrical (not shown here). Combining both posteriors using the rule (2) leads to a distribution shown in Figure 3c. This distribution is incorrect: it suggests the opponent could have slipped through the robots' perceptual fields. The correct posterior (for robots with perfect detection) is shown in Figure 3d. Such a posterior cannot be obtained by combining the two robots' momentary beliefs. This example illustrates why the simple form (2) is inapplicable to dynamic environments.

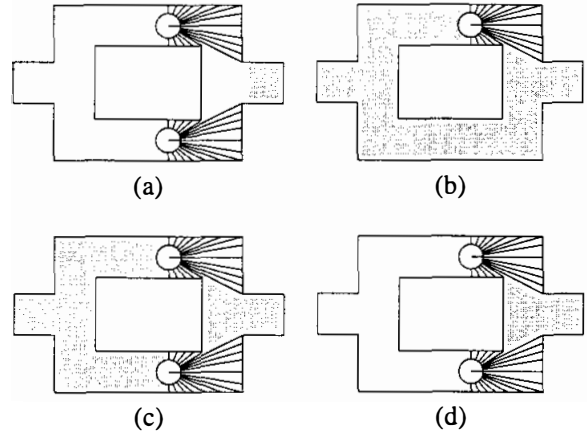

**Figure 3**: Illustration of why combining posterior states of individual platforms yields false results in dynamic environments. See text for details.

The obvious solution to this problem is to estimate the posterior over the entire state sequence $x_{0:t}$, not just the momentary state $x_t$. In particular, the following factorization applies (with essentially the same derivation as in Eq. 2):

$$p(x_{0:t} \mid d_{0:t}^1, \ldots, d_{0:t}^N) \propto p(x_{0:t}) \prod_{i=1}^N \frac{p(x_{0:t} \mid d_{0:t}^i)}{p(x_{0:t})}$$

Unfortunately, the dimension of the evidence $\frac{p(x_{0:t} \mid d_{0:t}^i)}{p(x_{0:t})}$ grows over time. So, we can no longer communicate all of our evidence in a fixed-size message.

## 3 Approximating the Bayes Filter

The above argument demonstrates that, in general, we cannot combine two pieces of evidence into a compact representation unless they refer to the same state of the world. In a dynamic environment, that means we can only combine two observations if they were gathered simultaneously; for example, we could combine sensor readings taken by two different agents on the same time step.

Depending on the details of our observation model, combining simultaneously-gathered evidence may or may not save any bandwidth. For example, in a $d$-dimensional linear-Gaussian model such as the one studied by (Nettleton et al., 2000), each observation takes $O(d)$ space to store. We can combine arbitrarily many observations into a $d \times d$ evidence matrix, but doing so only saves space if we have $\Omega(d)$ separate observations. Usually sensor platforms are scarce compared to landmarks, so it is not worth combining evidence.

In our target application of laser tag, it is even harder to take advantage of the ability to combine simultaneous observations: our observations are laser range scans of a small portion of a map. We can combine arbitrarily many observations into a data structure the size of a map of the world, but this process will not save us any space unless we have enough agents that their visible regions overlap and tile the entire map.

Since we cannot profitably combine evidence, and since we cannot afford to communicate all evidence, our algorithm must instead decide which evidence is worth sending. The remainder of this section describes how we do so.

### 3.1 Selective communication

The key idea of our approach is to selectively communicate only the most informative sensor measurements. More specifically, our approach approximates the desired posterior as follows:

$$p(x_{0:t} \mid d^1_{0:t},\ldots,d^N_{0:t}) \approx p(x_{0:t} \mid d^1_{0:t}, \delta^2_{0:t},\ldots, \delta^N_{0:t}) \quad (3)$$

Here we assume, for notational convenience, that the platform of interest has index $i = 1$; the data $d^1_{0:t}$ in (3) is hence the local data acquired by platform $i = 1$. Each $\delta^j_{0:t}$ with $j \neq i$, however, is a *subset* of the data acquired by platform $j$:

$$\delta^j_{0:t} \subseteq d^j_{0:t} \quad (4)$$

Thus, (3) is an approximation. This approximation is justified when sensor data is locally redundant, as is the case in our surveillance application. In this and many other scenarios, a small number of measurements suffices to obtain an accurate approximation.

The crux to the efficiency of our approach, thus, lies in the appropriate choice of the set $\delta^j_{0:t}$. A simple choice would be to communicate every $k$-th measurement, for an appropriate value of $k$. However, such an approach would not be very specific; it would risk communicating uninformative sensor measurements while omitting ones that are more informative. In what follows, we will describe a communication scheme that leads to the exchange of approximately maximally informative sensor measurements.

### 3.2 Selecting Informative Measurements for Communication

The key contribution of our approach is a protocol that enables platforms to selectively communicate measurements that are maximally informative. The *information* of a measurement is given by its relative entropy. Let $d^j_\tau$ be a measurement in robot's $j$ local database, and write $D^1 = (d^1_{0:t}, \delta^2_{0:t},\ldots, \delta^N_{0:t})$ for the data available to robot 1. Then the information of $d^j_\tau$ relative to the 1st robot's local belief is given by

$$q(d^j_\tau) = D_{\text{KL}}\left(p(x_{0:t} \mid D^1) \parallel p(x_{0:t} \mid D^1, d^j_\tau)\right) \quad (5)$$

Here $D_{\text{KL}}$ denotes the KL divergence. The most informative measurement is obtained by maximizing this expression: $\bar{d}^j_\tau = \text{argmax}_\tau q(d^j_\tau)$. For the $j$-th robot to identify the most informative measurement requires knowledge of another robot's posterior belief. Thus, our approach requires the communication of the belief of a robot. This belief is a "query," and the "answer" is the most informative sensor measurement in the data base of another entity.

Unfortunately, communicating the belief can be prohibitively expensive. In particular, the belief comprises all past states, not just the most recent one. On this basis alone, it might appear that our selective information scheme is inefficient. It merely shifts communication overhead from the exchange of information to the exchange of queries. In the next section, however, we will describe an efficient implementation that communicates highly compact approximate posterior beliefs.

## 4 The Decentralized Particle Filter

### 4.1 Basic Algorithm

Our approach represents each platform's posterior belief by a local particle filter (Doucet et al., 2001; Liu & Chen, 1998). The motivation for using particle filters is two-fold. For one, particle filters can represent almost arbitrary posterior distributions; they are certainly well-suited to accommodate the types of uncertainty that arise in the distributed surveillance scenario. More importantly, particle filters estimate posteriors over entire paths, not just the current state. Put differently, each particle can be thought of as an entire history or trajectory, and the set of all particles represents an approximation of the posterior over trajectories. This well-known property of particle filters (which is *not* shared by Kalman filters) makes them well-suited for the type of posteriors required by our selective communication scheme.

In the particle filter, an agent's posterior $p(x_{0:t} \mid D^1)$ is represented by a set of weighted samples or particles. We will write $x^i_{0:t}$ for the $i$th particle in this set, and $w^i_t$ for its weight.

We need to implement two different operations for our particle filter: incorporation of new evidence, and propagation from time $t$ to time $t + 1$.

If agent 1 finds out a new measurement $d_\tau^j$ taken by agent $j$ at time $\tau$, we can incorporate this evidence by setting

$$\begin{aligned} w_t^i &\leftarrow w_t^i p(d_\tau^j \mid x_{0:t}^i) \\ &= w_t^i p(d_\tau^j \mid x_\tau^i) \end{aligned} \quad (6)$$

for all $i$. This is exactly the standard particle filter observation update, with a simplification because $x_\tau$ d-separates $d_\tau^j$ from all other variables.

To propagate forward a time step, we pick a particle $x_{0:t}^i$ at random according to the weights $w_t^i / \sum_i w_t^i$. Then we sample a new particle $x'_{0:t+1}$ for time $t + 1$ according to

$$P(x'_{0:t+1} \mid x_{0:t}) = \begin{cases} p(x'_{t+1} \mid x'_t) & \text{if } x'_{0:t} = x_{0:t} \\ 0 & \text{otherwise} \end{cases} \quad (7)$$

### 4.2 Communication

To implement the query step of our communication algorithm, we could send the set of particles that characterizes a robot's local belief. Since each particle contains a sequence of past states, this message may still be prohibitively expensive. Thus, our approach employs a final approximation step: instead of communicating the full particle set, we only send a very small subset of particles. In particular, a query is composed of $M$ randomly selected particles, where each particle contains an entire state trajectory (within a fixed history window).

The robot that is being queried then searches its local database for the sensor measurement that is most informative, measured by the KL divergence criterion described above. It may choose one of its own measurements, or it may pass on a measurement which it has received from a third robot. The only restriction we impose is that it must send a relatively current observation, that is, one from some time $\tau \geq t - \Delta$. (This condition is not overly restrictive: old measurements are frequently uninformative anyway. And, it saves storage and computation, since states and observations older than $t - \Delta$ may be discarded.)

While our selection algorithm is somewhat sensitive to outliers (because they affect the posterior more than other measurements do), we found it to be extremely effective in our surveillance application.

### 4.3 Re-simulation

Unfortunately, the procedure as described so far sometimes results in large variance in the particles' importance weights. This variance can cause some importance weights to drop to zero, reducing the effective number of particles in our filter and thereby wasting computation. There are two reasons for this behavior: first, because communication between agents our particle filter receives more observation information than it would with just one agent's sensors. It is well known that particle filters behave poorly when observations are very informative compared to the proposal distribution (Thrun & Fox, 2000). Second, if our observation comes from several steps in the past ($\tau \ll t$), many of our particles will have overlapping histories at time $\tau$. That means that we will have fewer distinct importance weights, and so the normalization step will introduce extra variance.

In order to reduce the variance of our importance weights, our approach periodically re-simulates each agent's particle filter. That is, it erases part of the filter's history and then runs it forward again to the current time incorporating all collected observations at the appropriate times. Our current implementation re-simulates from time $\tau$ whenever we receive an observation for some time $\tau < t$. Re-simulation is related to well-known techniques for improving sample diversity in particle filters; for more detail see, *e.g.*, (van der Merwe et al., 2001).

## 5 Experimental Results

A systematic series of experiments was performed to evaluate the scaling properties of our approach. Experiments were performed both in simulation and using physical robots. Simulation enabled us to test the ability of our algorithm to scale to large numbers of robots, whereas the physical testbed provided us with insights into the practical effectiveness of our approach in the context of an actual robot system. In both cases, the scenario involved a distributed tracking task motivated by the laser tag problem. A team of robots was tasked to estimate a posterior distribution over an opponent's location from laser range data. In contrast to classical distributed tracking problems (Dolan et al., 1999), the laser tag domain is characterized by the pervasive *occlusion* of opponents. As a result, posterior distributions are highly non-Gaussian and multi-modal.

### 5.1 Physical Robot Results

In a first series of experiments, we collected data from a team of four robots in an indoor arena of size 25 by 20 meters, shown in Figure 4(a). The arena was outfitted with a number of obstacles, causing significant occlusion problems. A belief tracker for these robots was described in (Rosencrantz et al., 2003); however, the filter described in that publication is centralized. To evaluate the effectiveness of our communication scheme, we also implemented an alternative non-selective algorithm of communicating every $k$-th sensor measurement, for an appropriate value of $k$. This algorithm is a non-selective counterpart to our approach; it does not respond to queries, and it does not tailor its communication to the particular uncertainty faced by neighboring robots. The lack of a query phase reduces the total communication overhead, but at the expense of communicating measurements of inferior information value. Thus, by comparing our approach to this straw man approach we can evaluate the practical utility of selective information exchange.

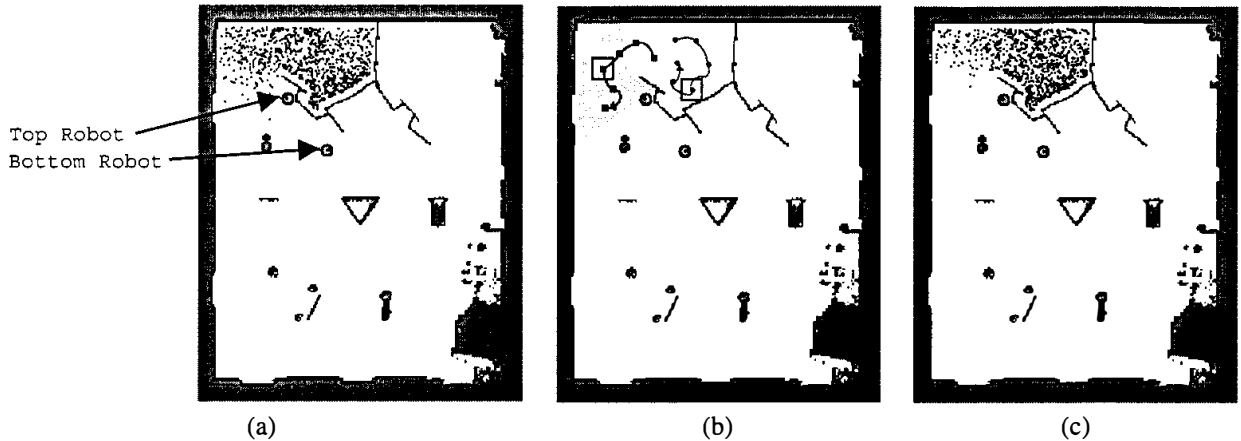

Figure 4: Illustration of the algorithm: Panel (a) shows the belief of the bottom robot with regards to the possible location of an opponent. Panel (b) illustrates a query communicated by the bottom robot. In response, the top robot determines that the measurement represented by the shaded area is most informative. After communicating this measurement back to the bottom robot, its posterior becomes the one shown in Panel (c). Note that (a) and (c) in this case are similar, this is because a relatively old observation was incorporated and re-simulation has allowed the the bottom robots distribution to spread again after that old observation was incorporated.

A simple example of our protocol in this domain is given in Figure 4. Two robots, illustrated by the two circular objects, are attempting to track an opponent (not shown here). Initially, the opponent is known to be in the top left are of the map. The local posterior of the bottom robot is shown in Figure 4a; it confines the opponent to the occluded are in the top left of the map. Unfortunately, the bottom robot does not survey the target area. A query, selected using our approach by the bottom robot, is shown in Figure 4b (somewhat idealized for visual clarity). Notice that this query consists of two paths, chosen from the bottom robot's set of particles. The shaded region represents the observation the top robot has chosen as most informative and the small squares show the time along each trajectory that the observation applies to. After communicating this measurement back to the bottom robot, the bottom robot's belief evolves to the one shown in Figure 4c. Notice that this new belief is similar to the one it started with in Figure 4a, this is because re-simulation has allowed the bottom robots posterior to spread after this new observation was incorporated. The reflects the fact that a given distribution can have a large number of potential pasts. In general we will need many observations to rule out all impossible histories.

To evaluate the performance of our approach systematically, we varied the rate at which the team members exchange information, from three times a second to once every three seconds. Each rate induces a different total communication bandwidth: in the non-selective scheme, the bandwidth is determined by the sensor measurements that are being broadcast through the system. In the selective scheme, the queries add additional overhead to the total bandwidth. Figure 5 shows the results of the comparison, for different values of the communication bandwidth. This graph is obtained by measuring the KL-divergence of the resulting posterior over $x_t$ from the posterior achieved by a particle filter with full communication (using Laplace smoothing over a fine-grained grid to interpolate between different discrete measurements). Each point is averaged over 6 runs and the error bars indicate the standard deviation at that point. This result shows that our algorithm outperforms the non-selective alternative as bandwidth becomes restricted. Put differently, our selective communication scheme leads to improved posterior estimates when compared to a non-selective alternative. All of these results were obtained using physical robot data, and they reflect the performance one could expect in an actual robotic laser tag setting.

### 5.2 Simulation Results

The physical testbed prohibits us from conducting experiments with larger number of platforms. This is because the number of robots available to us is limited. To investigate the scaling abilities of our approach, we developed a multi-robot simulator capable of handling very large numbers of robots.

Figure 6 shows the performance obtained with a simulation of 50 robots. Shown there are two graphs, one for our approach and one for the non-selective counterpart.Each graph is accompanied by curves characterizing 95% confidence intervals. As is easily seen, our algorithm outper-forms the naive algorithm. It maintains a satisfactorily low divergence at lower bandwidths. The margin of improvement is more significant than in the physical robot experiment, which we attribute to the superior scaling ability to large number of robots. In conclusion, our approach consistently generates excellent state estimation results in dynamic environments using a large number of robots.

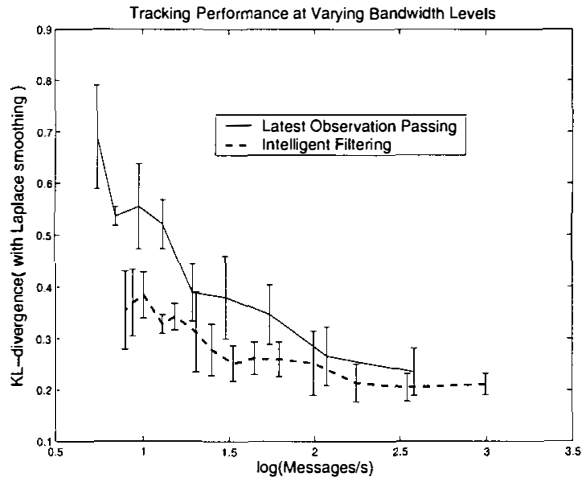

Figure 5: A comparison of our method with an alternative where robots send their most recent observation at varying frequencies. Our method outperforms this alternative, maintaining distributions closer to the truth.

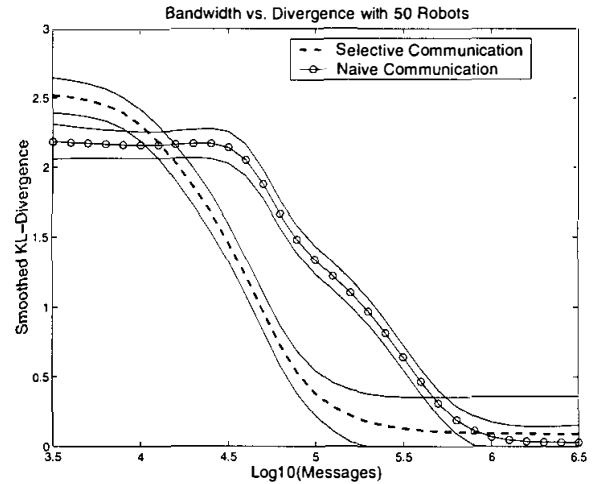

Figure 6: A comparison of our method with an alternative where robots send their most recent observation. Our method outperforms this alternative, maintaining distributions closer to the truth for a range of bandwidths.

## 6 Discussion

We have presented a scalable and robust distributed particle filter for dynamic environments. In contrast to previous research on decentralized filtering, our approach can cope with non-Gaussian posteriors and with dynamic environments. The key technical innovation is a selective communication scheme that enable individual platforms to communicate the most informative piece of information to other entities. Our approach utilizes particle filters, for which this idea can be implemented efficiently, and which are capable of representing non-Gaussian posteriors.

Systematic experimental results were conducted in the context of a distributed surveillance scenario, motivated by a robotic laser tag testbed presently under development at CMU. The results, using both simulation and physical robots, illustrate the effectiveness of our approach when compared to a straw-man approach that communicates information in a non-selective way. Simulations with up to 50 robots show superior performance of our approach.

There remain several open questions that warrant future research. Chief of those open questions is a more informed strategy for composing queries than random selection of particles. Clearly, certain particles are more suited for querying neighbors than others, and by taking advantage of this the overall performance may improve. The present algorithm for generating answers is also limited, in that it may favor outlier measurements on the sheer basis that they surprise. We are presently seeking ways to assess the information of measurements relative to the true (but unknown) posterior, to avoid communicating high-noise measurements. However, in our implementation this is not a problem, largely because of the low degree of noise in laser range measurements.

Another direction for future work is the intelligent selection of which agent to query. In our implementation robots chose which peer to communicate with at random from their $k$ nearest neighbors. This is a natural choice because many networks have the property that only nearby agents can communicate. However, this is not the only possibility, and choosing carefully when the underlying network can offer more flexibility deserves attention.

Another open question is how to improve re-simulation. The re-simulation step is the most computation-intensive part of our algorithm; so, in future work we plan to experiment with ways to avoid re-simulating as often. Possibilities include re-simulating only when too many importance weights get small, or re-simulating only some of the particles in our filter. A final opportunity is to extend our approach to one in which entities consider their control objective in their composition of queries. For example, in the laser tag domain, robots seek opponents, and the choice of particles in a query could incorporate the location of alleged opponents. To that end, we believe that our decentralized information fusion approach opens new opportunities for the decentralized control of teams of mobile robots.

Regardless of these limitations, we believe our that our approach is unique in its ability to accommodate non-Gaussian posteriors in a strictly decentralized fashion. Furthermore, we believe that it is applicable to a number of distributed state estimation problems in the context of ubiquitous computing.